\documentclass[letterpaper, 10 pt, conference]{ieeeconf}  

\IEEEoverridecommandlockouts               
\usepackage{amsmath,amssymb,hyperref,url,mathrsfs,array,graphicx,bm,balance,xcolor,subfig}
\usepackage[T1]{fontenc}

\newcommand{\figref}[1]{Fig.~\ref{#1}}
\newcommand{\secref}[1]{Sec.~\ref{#1}}

\title{\LARGE \bf
Safety-critical Motion Planning for Collaborative Legged Loco-Manipulation over Discrete Terrain}

\author{Mohsen Sombolestan and Quan Nguyen
\thanks{M. Sombolestan and Q. Nguyen are with the Department of Aerospace and Mechanical Engineering, University of Southern California, Los Angeles, CA 90089, email: {\tt somboles@usc.edu, quann@usc.edu}.}}

\begin{document}

\maketitle
\begin{abstract}
    As legged robots are deployed in industrial and autonomous construction tasks requiring collaborative manipulation, they must handle object manipulation while maintaining stable locomotion. The challenge intensifies in real-world environments, where they should traverse discrete terrain, avoid obstacles, and coordinate with other robots for safe loco-manipulation. This work addresses safe motion planning for collaborative manipulation of an unknown payload on discrete terrain while avoiding obstacles. Our approach uses two sets of model predictive controllers (MPCs) as motion planners: a global MPC generates a safe trajectory for the team with obstacle avoidance, while decentralized MPCs for each robot ensure safe footholds on discrete terrain as they follow the global trajectory. A model reference adaptive whole-body controller (MRA-WBC) then tracks the desired path, compensating for model uncertainties from the unknown payload. We validated our method in simulation and hardware on a team of Unitree robots. The results demonstrate that our approach successfully guides the team through obstacle courses, requiring planar positioning and height adjustments, and all happening on discrete terrain such as stepping stones.
    \
\end{abstract}
\section{Introduction} \label{sec: Introduction}

Legged robots are recognized for their rapid movement and ease of maneuvering due to their flexible locomotion capabilities. The progress in model predictive control (MPC) for legged robots \cite{DiCarlo2018, Li2021} has enabled the development of real-time control systems capable of executing various walking gaits. Most studies on quadruped robots have focused on locomotion \cite{Focchi2017, Bledt2018} and loco-manipulation by individual robots \cite{Chiu2022, Sleiman2021, Zimmermann2021, Rigo2023, Wolfslag2020OptimisationRobots, Ferrolho2023RoLoMa:Arms}. These approaches also address issues with significant uncertainties in the robot model \cite{Sombolestan2024, Sombolestan2021} and handling objects with unknown properties \cite{Sombolestan2023b}. However, there is limited research on the collaboration among multiple quadruped robots. A collaboration between multiple quadruped robots becomes highly advantageous when a single robot or in-hand manipulation is insufficient for handling large, heavy objects. The challenge becomes more significant in real-world situations, such as industrial factories and last-mile delivery operations, where the team of robots must safely navigate discrete terrain—a capability unique to legged robots compared to other ground robots.

Collaborative object manipulation has been a subject of research since the early developments in robot manipulators \cite{Tarn1986COORDINATEDARMS, Khatib1988ObjectSystem} and mobile robots \cite{Khatib1996}. This direction has also been extended to legged robots. Researchers have investigated using multiple quadruped robots towing a load with cables toward a target while avoiding obstacles \cite{Yang2022b}. A recent trend involves using interconnected legged robots for collaborative manipulation with holonomically constant properties defining the configuration setup. For instance, \cite{Kim2023LayeredApproaches} developed both centralized and distributed MPCs as high-level planners, followed by a distributed whole-body tracking controller that makes two quadruped robots to carry a payload. 
They further expand the application to obstacle avoidance by integrating control barrier functions into the high-level planner \cite{Kim2023Safety-CriticalFunctions}.
Another study introduced a passive arm to facilitate collaboration between robots and between robots and humans for payload transportation \cite{Turrisi2024PACC:Control}. This research area has expanded to bipedal robots, where researchers designed a decentralized controller using reinforcement learning for multi-biped robot carriers, adaptable to varying numbers of robots \cite{Pandit2024LearningTransport}.

\begin{figure}[t!]
    \center	\includegraphics[width=1\linewidth]{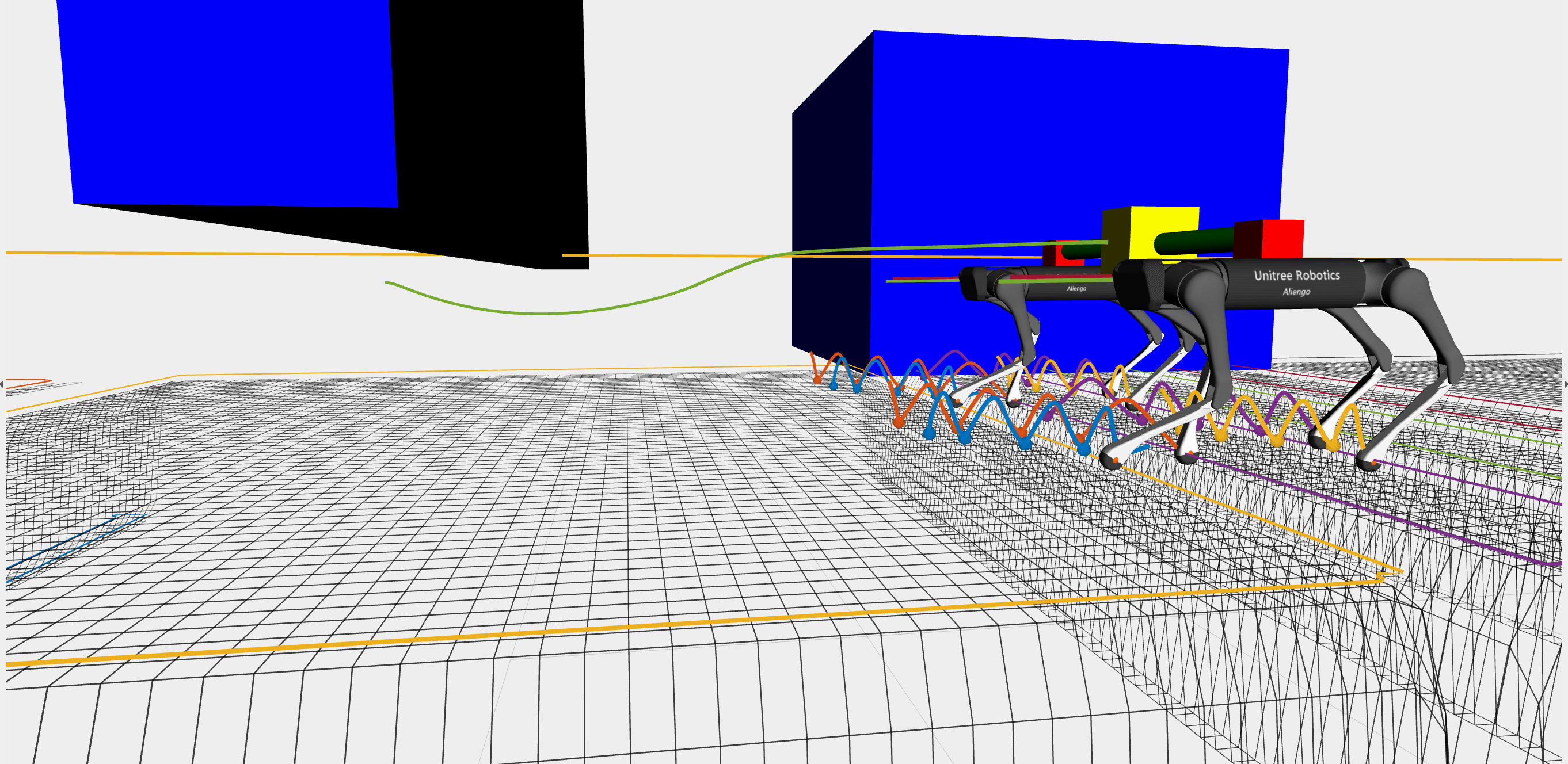}
    \caption{{\bfseries A team of quadrupedal robots transporting a payload}. The robots navigate a challenging obstacle course on discrete terrain, stepping on stones, crawling under the horizontal wall, and adjusting their height to pass over the vertical bar. More details in \url{https://youtu.be/MuJY9rYxTO4}.}
    \label{fig: interconnected_robots}
    \vspace{-0.6cm}
\end{figure}

\begin{figure*}[t!]
    \center	\includegraphics[width=0.91\linewidth]{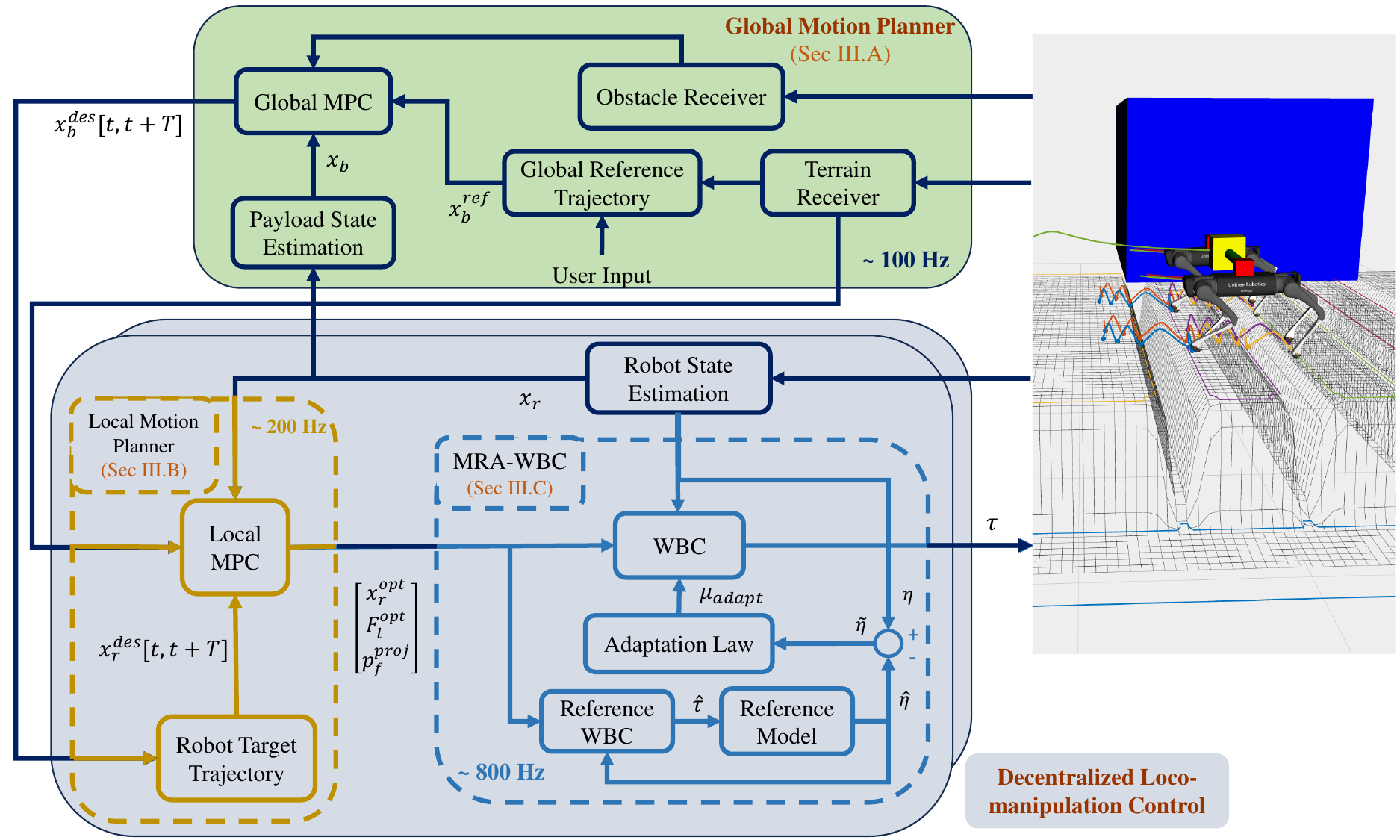}
    \caption{{\bfseries Diagram of the Proposed Method}. The global planner defines the team’s trajectory while avoiding obstacles. A decentralized loco-manipulation controller operates on each robot that uses local MPC to plan future footholds, followed by MRA-WBC to track the trajectory and manage uncertainties from the unknown payload.}
    \label{fig: block_diagram}
    \vspace{-0.5cm}
\end{figure*}

However, in all the mentioned research, payload transportation is primarily demonstrated in safe conditions on flat terrain, with minimal or no obstacles, and with prior knowledge of the payload's properties, which does not accurately represent real-world scenarios. Real-world applications require a safety-critical motion planner that not only optimizes the robots' trajectory to navigate environments with many obstacles safely but also ensures stable locomotion by selecting safe foothold positions on discrete terrain. The high-dimensional state space, nonlinear dynamics, and the need for coordination and communication among the robots further complicate this challenge.

In this paper, we propose a hierarchical approach for safety-critical motion planning of collaborative legged loco-manipulation over challenging terrain (as shown in \figref{fig: interconnected_robots}). This hierarchical control structure (illustrated in \figref{fig: block_diagram})—covering obstacle avoidance, safe foothold planner, and adaptive whole-body control—effectively addresses high-dimensional state spaces and uncertainties. The main contributions of this paper are as follows:
    \begin{itemize}
        \item We introduce a global MPC-based motion planner that generates an optimal and safe trajectory for the entire team, considering obstacle avoidance and terrain height.
        \item The global trajectory is then used as a reference by each robot's decentralized MPC, which acts as a local planner to generate optimal trajectories while ensuring safe footstep placement on discrete terrain.
        \item The decentralized MPC is followed by a model reference adaptive whole-body controller (MRA-WBC), enhancing robustness and ensuring accurate trajectory tracking despite uncertainties from the payload.
        \item Our approach is validated through simulations and experimental results with quadrupedal robots in various scenarios, including discrete and challenging terrains, unknown payloads, and 3D collision avoidance.
    \end{itemize}

The paper is structured as follows: \secref{sec: overview} outlines the proposed control system, and \secref{sec: Method} provides details on the global planner design and decentralized loco-manipulation controller. Simulation and hardware results are presented in \secref{sec: results}, followed by concluding remarks in \secref{sec: conclusion}.
\section{System Overview} \label{sec: overview}

This section briefly overviews our proposed safety-critical motion planning approach for collaborative loco-manipulation. In this setup, two robots are connected by a rigid rod and share the task of carrying an unknown payload. The rod has two rotational degrees of freedom at each connection point, allowing it to rotate freely and enabling the team to coordinate the payload's height and orientation.

Our method, shown in \figref{fig: block_diagram}, consists of two layers. The global motion planner uses MPC to generate an optimized team trajectory based on the user-defined target and terrain height map, considering collision avoidance. Then, a decentralized MPC on each robot adjusts the global trajectory to create optimal local trajectories, accounting for the future safe footstep locations. Finally, an MRA-WBC generates joint torques to track the desired trajectory despite payload uncertainties, formulated as a quadratic program (QP) minimizing tracking errors while guaranteeing stability through a control Lyapunov function (CLF).

The hierarchical control framework simplifies collaborative loco-manipulation on challenging terrain by separating the global and local MPCs. The global MPC handles obstacle avoidance, while the local MPC ensures stable footholds, prioritizing locomotion safety over global trajectory optimization. Adaptation is implemented in the whole-body controller as the last step in the control pipeline to ensure accurate tracking despite payload uncertainties. Introducing adaptation at the planning level would add adaptive-CLF constraints \cite{Minniti2021}, making the MPC problems harder to solve. However, using nominal payload properties during planning simplifies the process, while the MRA-WBC adapts to uncertainties afterward.

\section{Proposed Method} \label{sec: Method}
In this section, we will provide a detailed explanation of each component of our control system pipeline, as shown in \figref{fig: block_diagram}. We will break down the structure and function of each element, highlighting how they contribute to achieving the goal of collaborative loco-manipulation on discrete terrain.

\subsection{Global Motion Planner}
The main objective of the global motion planner is to create a safe and optimized path to the user's target location while considering obstacle avoidance. Initially, a reference trajectory $\bm{x}_b^{ref}$ is generated based on the user's specified location, considering the height map so that each reference point remains above the terrain, ensuring that robots can navigate uneven surfaces. The global motion planner, modeled as an MPC problem, optimizes the team trajectory to follow the reference trajectory while guaranteeing safety and collision avoidance.
The state vector comprises the payload's position, quaternion, linear velocity, and angular velocity, represented as \( \bm{x}_b = [\bm{p}_b^T, \bm{q}_b^T, \bm{v}_b^T, \bm{\omega}_b^T]^T \). The input vector \( \bm{u}_b = [\bm{F}_b^T, \bm{\tau}_b^T]^T \) includes the forces and moments applied to the payload. The MPC problem can be formulated as follows:
\vspace{-0.3cm}
\begin{subequations}
    \begin{align}
         & \underset{\bm{u}_b(\cdot)}{\text{minimize}} &  & S_b(\bm{x}_b(T)) + \int_0^T l_b(\bm{x}_b(t), \bm{u}_b(t), t)  dt              \\
         & \text{subject to:}                          &  & \dot{\bm{x}}_b(t) = f(\bm{x}_b) + g(\bm{x}_b) \bm{u}_b \label{eq: dynamic_eq} \\
         &                                             &  & \bm{h}(\bm{x}_b, \bm{u}_b) \geq \bm{0} \label{eq: inequality}
    \end{align}
\end{subequations}
where \( S_b(.) \) represents the final cost, and \( l_b(.) \) denotes the intermediate cost over a time horizon \( T \), subject to both dynamic \eqref{eq: dynamic_eq} and inequality constraints \eqref{eq: inequality}. The payload states, attached at the center of the connecting rod, can be determined from the states of both robots based on the offset between each robot's base and the rod connection point.

Note that the decentralized loco-manipulation controller receives only the optimized target trajectory of the payload, \( \bm{x}_b^{\text{des}}[t, t+T] \), from the global motion planner for a future horizon \( T \) (see \figref{fig: block_diagram}). While the control input \( \bm{u}_b \) is not used in the decentralized controller, considering cost and bounds for \( \bm{u}_b \) ensures the MPC generates a realistic trajectory. This is why MPC is preferred over traditional path-planning algorithms for the global motion planner.

\subsubsection{Cost function}
The objective of the cost function is to minimize the deviation of payload state $\bm{x}_b$ from the reference trajectory $\bm{x}_b^{ref}$ while minimizing the input  \( \bm{u}_b \). We propose the following quadratic cost function:
\begin{align} \label{eq: cost cont}
    l_b(\bm{x}_b(t), \bm{u}_b(t), t) = & \frac{1}{2}  \|\bm{x}_{b}(t) - \bm{x}_{b}^{\text{ref}}(t)\|_{\bm{Q}_{x_b}}
    + \frac{1}{2} \|\bm{u}_b(t)\|_{\bm{R}_u}
    \nonumber                                                                                                       \\
    S_b(\bm{x}_b(T)) =                 & \frac{1}{2}  \|\bm{x}_{b}(T) - \bm{x}_{b}^{\text{ref}}(T)\|_{\bm{Q}_{f}}
\end{align}
where $\|\mathbf{a}\|_{\bm{Q}}$ denotes the weighted norm $\mathbf{a}^T \bm{Q} \mathbf{a}$ and $\bm{Q}_f$, $\bm{Q}_{x_b}$, and $\bm{R}_u$ are defining the weight matrices.

\subsubsection{Dynamic equation}
The payload's dynamic equation, including the connecting rod, is expressed as:
\begin{equation}
    \dot{\bm{x}}_b =
    \begin{bmatrix}
        \bm{v}_b                                   \\
        \frac{1}{2} \bm{q}_b \otimes \bm{\omega}_b \\
        \bm{0}                                     \\
        -\bm{I}_G^{-1} \bm{\omega}_b \times \bm{\omega}_b
    \end{bmatrix}  +
    \begin{bmatrix}
        \bm{0}               & \bm{0}        \\
        \bm{0}               & \bm{0}        \\
        \frac{1}{m} \bm{I}_3 & \bm{0}        \\
        \bm{0}               & \bm{I}_G^{-1}
    \end{bmatrix}
    \begin{bmatrix}
        \bm{F}_b \\
        \bm{\tau}_b
    \end{bmatrix}
\end{equation}
where \( m \) and \( \bm{I}_G \) denote the combined mass and moment of inertia, and \( \otimes \) represents quaternion multiplication.

\subsubsection{Inequality constraints}
To ensure safety and obstacle avoidance, we define the following barrier functions:
\begin{subequations}
    \begin{align}
        B_{j}^b(\bm{x}_b) \stackrel{\Delta}{=}     & \|\bm{\mathcal{O}}_j - \bm{p}_b \| - \mathcal{R}_{j,b} \geq 0
        \label{eq: payload_constraints}                                                                                                                \\
        B_{j}^{r,i}(\bm{x}_b) \stackrel{\Delta}{=} & \|\bm{\mathcal{O}}_j - \bm{p}_{r,i} \| - \mathcal{R}_{j,r,i} \geq 0 \label{eq: robot_constraints}
    \end{align}
\end{subequations}
where $\bm{p}_{r, i}$ represents the position of robot $i$ and \(\bm{\mathcal{O}}_j\) denotes the position of obstacle $j$. \(\mathcal{R}_{j,b}\) is the obstacle's barrier radius, considering the size of the payload, and \(\mathcal{R}_{j,r,i}\) accounts for the size of robot $i$. Note that the robot's position can be derived from the payload's position, and vice versa. Therefore, $\bm{p}_{r, i}$ is a function of the payload state, $\bm{x}_b$.

The current setup for our collaborative manipulation allows us to coordinate the payload in three dimensions. Adjusting the payload orientation and the connecting rod can be achieved by modifying each robot's 3D position. The robots can move freely in the x-y plane, but there are constraints on each robot's height that need to be considered within the MPC problem. The robot's height can be determined based on the payload position and orientation as follows:
\begin{align}
    \begin{bmatrix}
        p^x_{r,i} \\
        p^y_{r,i} \\
        p^z_{r,i}
    \end{bmatrix} = \bm{p}_b + \bm{R}
    \begin{bmatrix}
        0       \\
        \pm l/2 \\
        0
    \end{bmatrix} + \bm{c}_i
\end{align}
where $\bm{R}$ is the rotation matrix of the rod, $l$ is the rod's length, with $\pm$ indicating the side of the rod to which the robot is attached, and $\bm{c}_i$ is the constant offset from the rod’s connecting point to the robot's base. Thus, we impose the allowable range \((h_{min}, h_{max})\) on each robot's optimized target height using the following constraint:
\begin{align} \label{eq: height_constraint}
    h_{min} \leq p^z_{r,i} \leq h_{max}
\end{align}
An allowable range for the force and moment, represented by the input vector, can also be defined as follows:
\begin{align} \label{eq: input_constraint}
    \bm{u}_{min} \leq \bm{u}_b \leq \bm{u}_{max}
\end{align}

Together, equations \eqref{eq: payload_constraints}, \eqref{eq: robot_constraints}, \eqref{eq: height_constraint}, and \eqref{eq: input_constraint} will establish the inequality constraints \eqref{eq: inequality} in our MPC formulation.

\subsection{Local Motion Planner}
As part of the decentralized loco-manipulation controller, the local motion planner uses an MPC problem to ensure each robot follows the optimized target trajectory \( \bm{x}_b^{\text{des}}[t, t+T] \) from the global planner. This local MPC considers safe foothold placement for each leg on discrete terrain, adjusting the target trajectory to maintain stable locomotion.

The full-body nonlinear dynamics of the robot, including the floating base and leg dynamics, can be expressed as:
\begin{equation} \label{eq: full-body dynamic}
    \bm{M}(\bm{\nu})\ddot{\bm{\nu}} + \bm{C}(\bm{\nu}, \dot{\bm{\nu}}) = \sum_{i=1}^{4} \bm{J}_i^T \bm{F}_{l,i} + \bm{B}^T \bm{\tau}
\end{equation}
with the generalized coordinates $\bm{\nu} = \left[\bm{p}_r^T, \bm{\Theta}^T, \bm{q}_j^T \right]^T$ and velocities $\dot{\bm{\nu}} = \left[\bm{v}_r^T, \bm{\omega}_r^T, \dot{\bm{q}}_j^T \right]^T$. Here, \( \bm{p}_r \) represents the robot's position, \( \bm{\Theta} \) denotes the orientation, \( \bm{v}_r \) and \( \bm{\omega}_r \) are translational and rotational velocities, respectively. \( \bm{q}_j, \dot{\bm{q}}_j \in \mathbb{R}^{12} \) represent the joint angles and velocities of all four legs, \( \bm{F}_l = [\bm{F}_{l,1}^T, \bm{F}_{l,2}^T, \bm{F}_{l,3}^T, \bm{F}_{l,4}^T]^T \in \mathbb{R}^{12} \) represents the ground reaction forces (GRFs) on the legs, \( \bm{\tau} \in \mathbb{R}^{12} \) is the joint torque vector, \( \bm{M} \in \mathbb{R}^{18 \times 18} \) is the inertia matrix, and \( \bm{C} \in \mathbb{R}^{18} \) represents the nonlinear terms. The contact Jacobian matrix for each leg is \( \bm{J}_i \in \mathbb{R}^{3 \times 18} \), and \( \bm{B} = [\bm{0}_{12 \times 6}, \bm{I}_{12 \times 12}] \in \mathbb{R}^{12 \times 18} \) is the selection matrix. We can only use the floating base dynamics to simplify the MPC formulation, as the leg inertia is negligible \cite{Grandia2022a}. The floating base dynamic is given by:
\begin{equation} \label{eq: floating-based dynamic}
    \bm{M}_B(\bm{\nu}_B)\ddot{\bm{\nu}}_B + \bm{C}_B(\bm{\nu}_B, \dot{\bm{\nu}}_B) = \sum_{i=1}^{4} \bm{J}_{B,i}^T \bm{F}_{l,i}
\end{equation}
with \( \bm{\nu}_B = \left[\bm{p}_r^T, \bm{\Theta}^T \right]^T \) and \( \dot{\bm{\nu}}_B = \left[\bm{v}_r^T, \bm{\omega}_r^T \right]^T \). The matrices \( \bm{M}_B \), \( \bm{C}_B \), and \( \bm{J}_{B,i} \) correspond to the floating base components of the full-body dynamics. However, for the whole-body controller discussed in \secref{sec: MRA-WBC}, the full-body dynamics must be used to compute the torque vector \( \bm{\tau} \).

The state and input vectors for the local MPC are:
\begin{equation}
    \bm{x}_r = \left[\bm{p}_r^T, \bm{\Theta}^T, \bm{v}_r^T, \bm{\omega}_r^T, \bm{q}_j^T \right]^T, \quad \bm{u}_r = \left[ \bm{F}_{l}^T, \dot{\bm{q}}_j^T \right]^T
\end{equation}
which considers the floating base and joint angles as the state variables, with GRFs and joint velocities as the inputs for the MPC. Since the target trajectory \( \bm{x}_b^{\text{des}}[t, t+T] \) specifies the payload desired states, we first need to compute the robot's target trajectory \( \bm{x}_r^{\text{des}}[t, t+T] \) based on the geometric relationship between the robots and the payload. Similar to the global MPC, the local MPC aims to minimize the deviation of the robot's state \( \bm{x}_r \) from the robot's target trajectory \( \bm{x}_r^{\text{des}}[t, t+T] \) with minimal effort, while also optimizing for future safe foothold positions on discrete terrain. The optimization problem can be formulated as follows:
\begin{subequations}
    \begin{align}
         & \underset{\bm{u}_r(\cdot)}{\text{minimize}} &  & S_r(\bm{x}_r(T)) + \int_0^T l_r(\bm{x}_r(t), \bm{u}_r(t), t)  dt                        \\
         & \text{subject to:}                          &  & \dot{\bm{x}}_r(t) = f_r(\bm{x}_r) + g_r(\bm{x}_r) \bm{u}_r \label{eq: dynamic_eq_local_mpc} \\
         &                                             &  & \bm{h}(\bm{x}_r, \bm{u}_r) \geq \bm{0} \label{eq: inequality_local_mpc}
    \end{align}
\end{subequations}
The dynamic equation \eqref{eq: dynamic_eq_local_mpc} is based on the floating base dynamic \eqref{eq: floating-based dynamic}.
For the footstep policy, we use a heuristic that places the nominal foothold beneath the hip at the midpoint of the contact phase \cite{Raibert1986}, with feedback based on the robot's base velocity:
\begin{equation}
    \bm{p}^{\text{nom}}_{f,i} = \bm{p}^{\text{nom}}_{f,i,\text{hip}} + \sqrt{\frac{p^{\text{des}}_{r,z}}{g}} (\bm{v}_{r} - \bm{v}^{\text{des}}_{r}),
\end{equation}
where \( \bm{p}^{\text{nom}}_{f,i} \) is the nominal foothold for leg \(i\), \( \bm{p}^{\text{nom}}_{f,i,\text{hip}} \) is the location beneath the hip, and \( g \) is gravity. Once set, footholds are projected onto the nearest plane based on the terrain receiver that is within kinematic limits to achieve a safe foothold on discrete terrain \cite{Grandia2022a}:
\begin{equation}
    \underset{\bm{p}^{\text{proj}}_{f,i} \in \Pi(\bm{p}^{\text{nom}}_{f,i})}{\text{minimize}}  \|\bm{p}^{\text{nom}}_{f,i} - \bm{p}^{\text{proj}}_{f,i}\| +  f_{\text{kin}}(\bm{p}^{\text{proj}}_{f,i}),
\end{equation}
where \( \bm{p}^{\text{proj}}_{f,i} \) is the projected foothold within the safe region, and \( f_{\text{kin}} \) is a kinematic penalty function. Further details on the cost function formulation and other locomotion constraints for the local MPC planner can be found in \cite{Grandia2022a}.

\subsection{Model Reference Adaptive Whole-body Control (MRA-WBC)} \label{sec: MRA-WBC}

From the local MPC, we get the optimized robot state \( \bm{x}^{\text{opt}}_{r} \), ground reaction forces \( \bm{F}^{\text{opt}}_{l} \), joints velocity \( \dot{\bm{q}}^{\text{opt}}_{j} \), and safe foothold positions \( \bm{p}^{\text{proj}}_{f,i} \). We then design a tracker controller to follow these values, as shown in \figref{fig: block_diagram}. The proposed MRA-WBC controller uses an adaptive technique to track the trajectory while handling payload uncertainties, guaranteeing stability and performance. Remember that applying the adaptive controller in the final control layer enhances robustness and accurate trajectory tracking.

\subsubsection{Feedback Linearization}
Let us Define the tracking error as:
\begin{equation}
    \bm{e} = \bm{\nu} - \bm{\nu}^{\text{opt}}, \quad \dot{\bm{e}} = \dot{\bm{\nu}} - \dot{\bm{\nu}}^{\text{opt}}
\end{equation}
where $\bm{\nu}^{\text{opt}}$ and $\dot{\bm{\nu}}^{\text{opt}}$ are derived from \( \bm{x}^{\text{opt}}_{r} \) and \( \dot{\bm{q}}^{\text{opt}}_{j} \). The state error vector is $\bm{\eta} = [\bm{e}^T, \dot{\bm{e}}^T]^T$. By applying state transformation and feedback linearization, the nonlinear dynamic \eqref{eq: full-body dynamic} is linearized into a system with the state error vector, relating the control input of the linear system to the nonlinear one. This method for force-based control is detailed in our previous work \cite{Sombolestan2024}. The linearized error dynamics are:
\begin{align} \label{eq: linearized_dynamics}
    \dot{\bm{\eta}}  = \bm{D} \bm{\eta} + \bm{G} \bm{\mu}, \quad
    \bm{D}           = \begin{bmatrix} \bm{0} & \bm{I} \\ \bm{0} & \bm{0} \end{bmatrix}, \quad \bm{G} = \begin{bmatrix} \bm{0} \\ \bm{I} \end{bmatrix}
\end{align}
where $\bm{\mu}$ is the control input for the linearized system.

Expanding equation \eqref{eq: linearized_dynamics}, the control input is derived as:
\begin{equation} \label{eq: control_input}
    \ddot{\bm{e}} = \ddot{\bm{\nu}} - \ddot{\bm{\nu}}_{d} = \bm{\mu}
\end{equation}
where $\ddot{\bm{\nu}}_{d}$ is the desired acceleration.
By applying a PD controller as the control input $\bm{\mu}$, we obtain:
\begin{equation}
    \bm{\mu} = \begin{bmatrix}
        -\bm{K}_P & -\bm{K}_D
    \end{bmatrix} \bm{\eta} , \quad
    \dot{\bm{\eta}} = \underbrace{\begin{bmatrix}
            \bm{0}    & \bm{I}    \\
            -\bm{K}_P & -\bm{K}_D
        \end{bmatrix}}_{\bm{A}_m} \bm{\eta}
\end{equation}
which is stable as long as $\bm{A}_m$ is Hurwitz.

\subsubsection{Control Lyapunov Function (CLF)}
To ensure the stability of the whole-body controller, we introduce a CLF \cite{Sontag1999}, which guarantees the stability of the closed-loop system and will later be utilized in the design of the QP for the whole-body controller. The CLF is defined as:
\begin{equation}
    V(\bm{\eta}) = \bm{\eta}^T \bm{P} \bm{\eta}
\end{equation}
where $\bm{P}$ is the solution to the Lyapunov equation $\bm{A}_m^T \bm{P} + \bm{P} \bm{A}_m = -\bm{Q}_L$, with $\bm{Q}_L$ being any symmetric positive definite matrix. The time derivative of the CLF is:
\begin{equation}
    \dot{V}(\bm{\eta}, \bm{\mu}) = \bm{\eta}^T (\bm{P} \bm{D} + \bm{D}^T \bm{P}) \bm{\eta} + 2 \bm{\eta}^T \bm{P} \bm{G} \bm{\mu}
\end{equation}
Hence, the CLF condition can be expressed as:
\begin{equation}
    \Psi_{clf} \stackrel{\Delta}{=} \dot{V}(\bm{\eta}, \bm{\mu}) + \lambda  V(\bm{\eta}) \leq 0
\end{equation}
where $\lambda$ is a positive scalar that ensures the CLF condition is satisfied.
\begin{figure*}[t!]
    \subfloat[]{\includegraphics[width=0.31\linewidth]{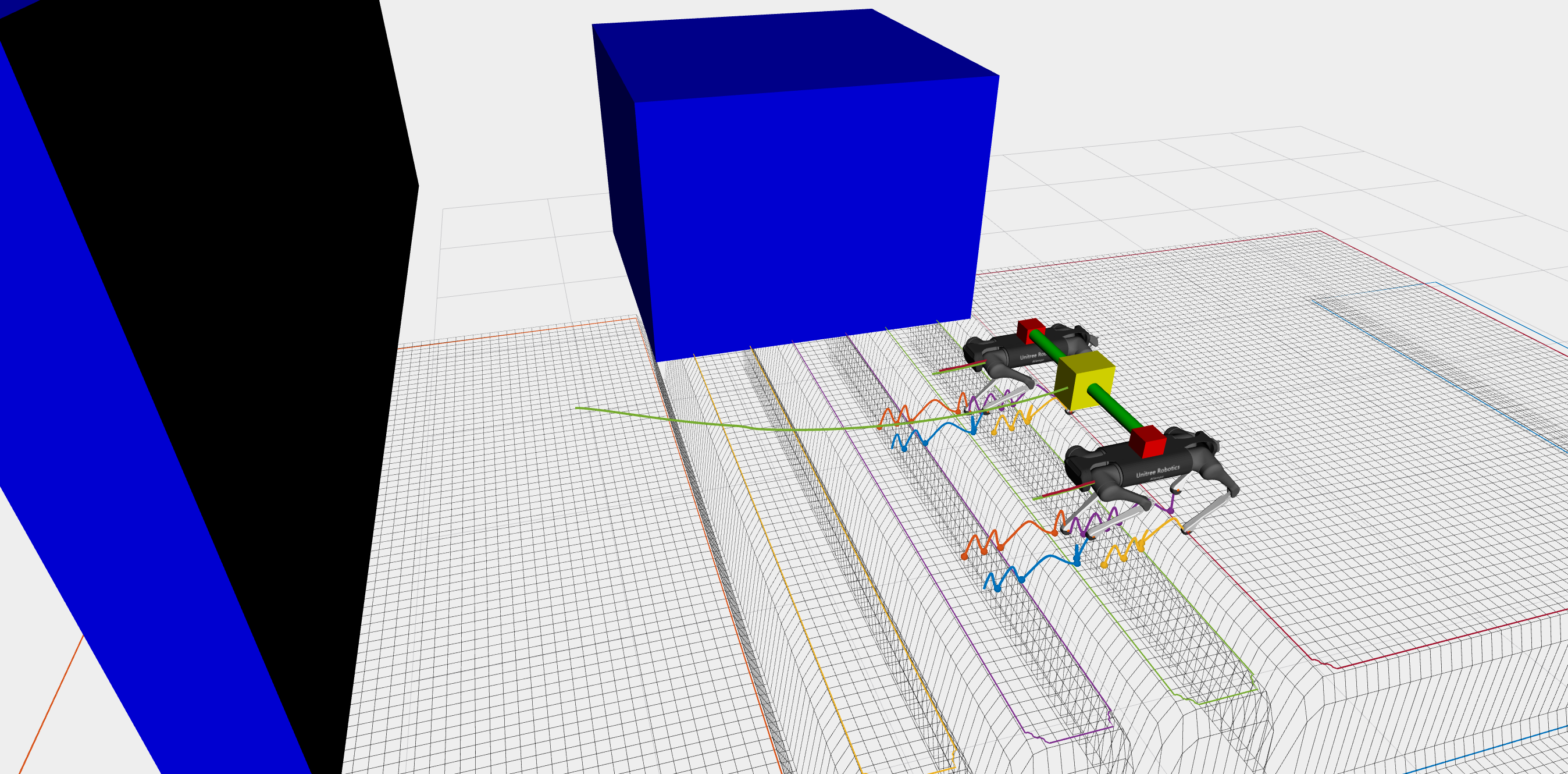} \label{fig: sim_fig_1}}
    \hfill
    \subfloat[]{\includegraphics[width=0.31\linewidth]{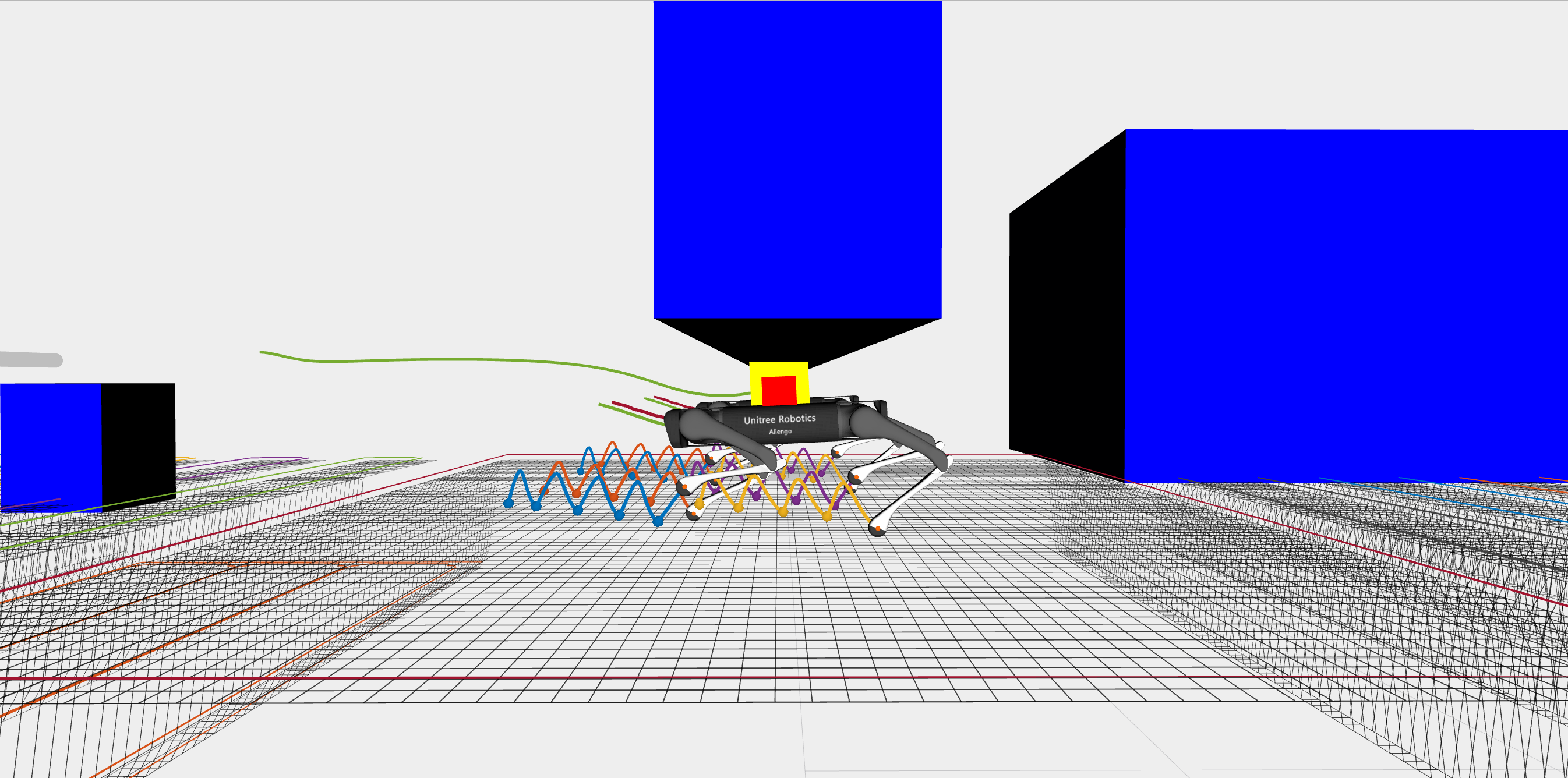} \label{fig: sim_fig_2}}
    \hfill
    \subfloat[]{\includegraphics[width=0.31\linewidth]{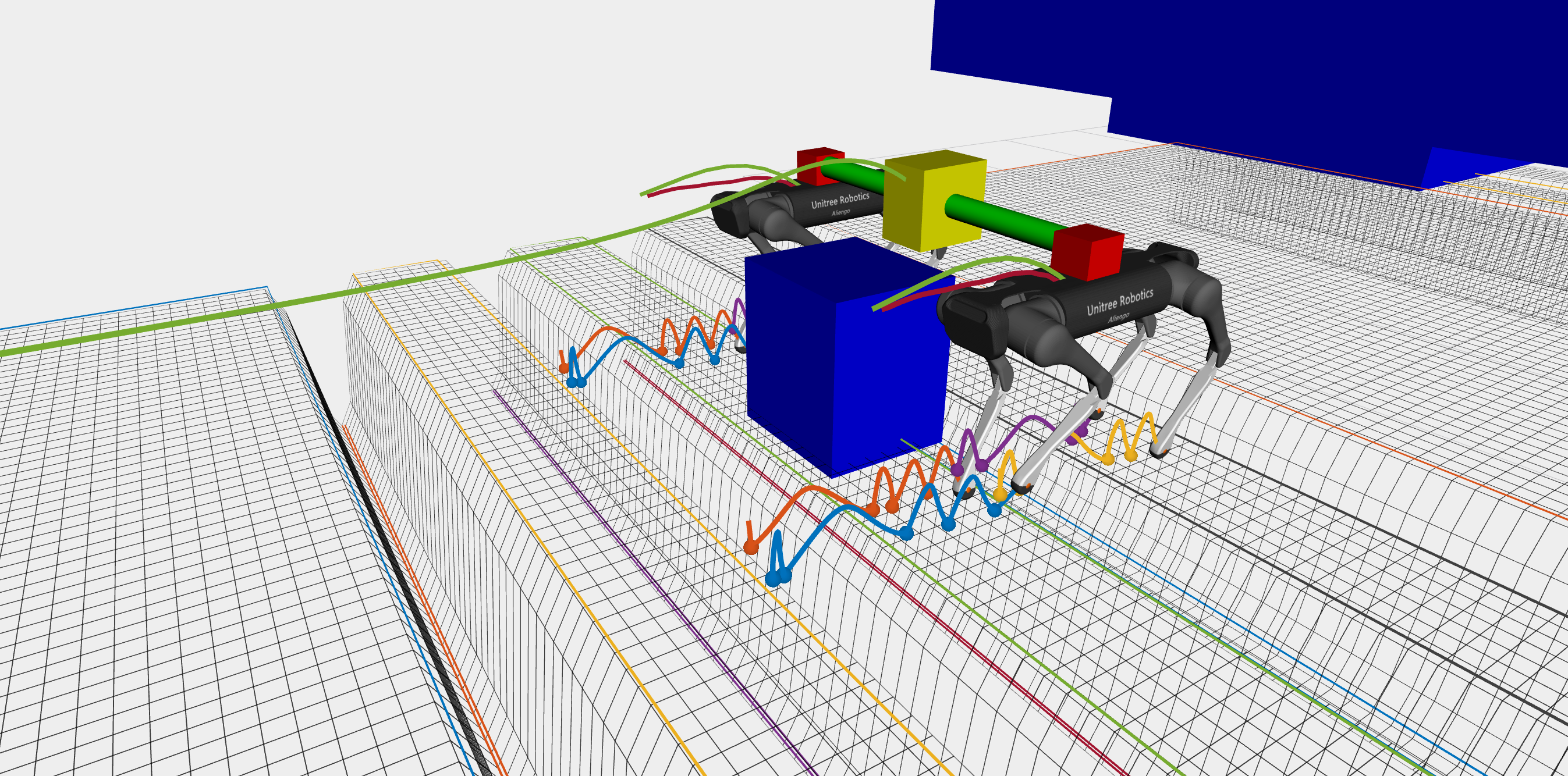} \label{fig: sim_fig_3}}
    \caption{\textbf{Robot team navigating an obstacle course with payload and disturbances.} The sequence shows (a) trajectory deviation to avoid an obstacle, (b) crawling under a low wall, and (c) adjusting height to clear a vertical bar. Through all these situations, the local MPC generates future safe foothold locations for stable locomotion on stepping stones.}
    \label{fig: sim_fig}
    \vspace{-0.5cm}
\end{figure*}

\subsubsection{Whole-body Controller (WBC)}
We can now design the whole-body controller as a QP, where the decision variables are the control input $\bm{\mu}$, joint torques $\bm{\tau}$, and the GRFs $\bm{F}_l$. The objective is to minimize the control input, the difference between the optimized GRFs $\bm{F}_{l}^{\text{opt}}$ and the actual GRFs $\bm{F}_l$, as well as to follow the desired foothold locations $\bm{p}_{f,i}^{\text{proj}}$. The QP is formulated as follows:
\begin{subequations} \label {eq: WBC}
    \begin{align}
         & \underset{\bm{\tau}, \bm{\mu}, \bm{F}_l}{\text{minimize}} &  & \|\bm{\mu}\|_{W_{\mu}} + \|\bm{F}_l - \bm{F}_{l}^{\text{opt}}\|_{W_{F}} \nonumber                                                             \\
         &                                                           &  & + \sum_{i=1}^{4} \|\bm{p}_{f,i} - \bm{p}_{f,i}^{\text{proj}} \|_{W_{p}}                                                                       \\
         & \text{subject to:}                                        &  & \bm{M}(\ddot{\bm{\nu}}_d + \bm{\mu}) + \bm{C} =  \sum_{i=1}^{4} \bm{J}_i^T \bm{F}_{l,i} + \bm{B}^T \bm{\tau}   \label{eq: dynamic_wbc} \\
         &                                                           &  & \bm{\tau}_{min} \leq \bm{\tau} \leq \bm{\tau}_{max}   \label{eq: torque_lim}                                                           \\
         &                                                           &  & \underline{\bm{d}} \leq \bm{C}_f \bm{F}_l \leq \bar{\bm{d}}      \label{eq: friction_cone}                                             \\
         &                                                           &  & \Psi_{clf} \leq 0
    \end{align}
\end{subequations}
Here, $W_{\mu}$, $W_{F}$, and $W_{p}$ are positive definite matrices defining the weights for each term in the cost function. Equation \eqref{eq: dynamic_wbc} represents the robot's full-body dynamics, with the control input substituted from equation \eqref{eq: control_input}. The constraints in \eqref{eq: torque_lim} enforce the torque limits, while \eqref{eq: friction_cone} imposes the friction cone constraints for the GRFs. Finally, the CLF condition $\Psi_{clf}$ ensures the controller's stability.

\subsubsection{Adaptive WBC}
Now, let us examine the impact of uncertainty arising from the unknown payload parameters on the robot’s dynamic parameters, such as in matrices $\bm{M}$ and $\bm{C}$. Assume that the unknown parameters introduce the following uncertainties in the dynamic parameters:
\begin{equation}
    \bar{\bm{M}} = \bm{M} + \tilde{\bm{M}}, \quad \bar{\bm{C}} = \bm{C} + \tilde{\bm{C}}
\end{equation}
where $\tilde{\bm{M}}$ and $\tilde{\bm{C}}$ represent the uncertainties in the inertia matrix and the nonlinear terms, respectively. These uncertainties affect the linearized dynamics \eqref{eq: linearized_dynamics}, as shown in \cite{Nguyen2015}:
\begin{equation} \label{eq: real_model}
    \dot{\bm{\eta}} = \bm{D} \bm{\eta} + \bm{G} \bm{\mu} + \bm{G} \bm{\theta}
\end{equation}
where $\bm{\theta}$ is the vector of uncertainties. The nonlinear uncertainty $\bm{\theta}$ is introduced through the uncertainties in $\tilde{\bm{M}}$ and $\tilde{\bm{C}}$, and is a function of $\bm{\theta}(\bm{\eta}, t)$ \cite{Nguyen2015}. For any time $t$, there exist functions $\bm{\alpha}(t)$ and $\bm{\beta}(t)$ such that \cite{L1_adaptive}:
\begin{equation}
    \bm{\theta}(\bm{\eta}, t) = \bm{\alpha}(t) \| \bm{\eta} \| + \bm{\beta}(t)
\end{equation}
Thus, we design a combined controller $\bm{\mu} = \bm{\mu}_{\text{\text{nom}}} + \bm{\mu}_{\text{adapt}}$, where $\bm{\mu}_{\text{\text{nom}}}$ is the nominal controller for reference tracking, and $\bm{\mu}_{\text{adapt}}$ is the adaptive controller signal compensating for the effect of uncertainty $\bm{\theta}$.

The structure of the proposed MRA-WBC is illustrated in \figref{fig: block_diagram}. A reference model is employed, using the robot’s dynamics but without uncertainties and utilizing a WBC as defined in \eqref{eq: WBC}, referred to as the \textit{reference WBC}. The reference model’s error state is denoted as $\hat{\bm{\eta}}$, and the error dynamics are given by:
\begin{equation} \label{eq: reference_error}
    \dot{\hat{\bm{\eta}}} = \bm{D} \hat{\bm{\eta}} + \bm{G} \hat{\bm{\mu}}_{\text{\text{nom}}} + \bm{G}( \hat{\bm{\theta}} + \bm{\mu}_{\text{adapt}})
\end{equation}
where $\hat{\bm{\mu}}_{\text{\text{nom}}}$ is the control input for the reference model, and $\hat{\bm{\theta}}$ is the estimate of the uncertainty vector. By employing the following adaptive controller, we can compensate for the estimate of uncertainties vector $\hat{\bm{\theta}}$:
\begin{equation} \label{eq: control_law}
    \bm{\mu}_{\text{adapt}} = - \hat{\bm{\theta}}
\end{equation}

The objective is for the real model $\bm{\eta}$ to track the reference model $\hat{\bm{\eta}}$. The difference between the real and reference models is defined as $\tilde{\bm{\eta}} = \hat{\bm{\eta}} - \bm{\eta}$. Considering equations \eqref{eq: real_model} and \eqref{eq: reference_error}, we obtain:
\begin{equation}
    \dot{\tilde{\bm{\eta}}} = \bm{D} \tilde{\bm{\eta}} + \bm{G} \tilde{\bm{\mu}}_{\text{\text{nom}}} + \bm{G} (\tilde{\bm{\alpha}} \| \bm{\eta} \| + \tilde{\bm{\beta}})
\end{equation}
with $\tilde{\bm{\mu}}_{\text{\text{nom}}} = \hat{\bm{\mu}}_{\text{\text{nom}}} - \bm{\mu}_{\text{\text{nom}}}$, $\tilde{\bm{\alpha}} = \hat{\bm{\alpha}} - \bm{\alpha}$, and $\tilde{\bm{\beta}} = \hat{\bm{\beta}} - \bm{\beta}$. Now, we can estimate $\bm{\theta}$ indirectly through $\bm{\alpha}$ and $\bm{\beta}$, or their estimates $\hat{\bm{\alpha}}$ and $\hat{\bm{\beta}}$, which are updated using the following adaptation laws based on projection operators \cite{Lavretsky2011}:
\begin{align}
    \label{eq: adap_law}
    \dot{\hat{\bm{\alpha}}} = \bm{\Gamma}\text{Proj}(\hat{\bm{\alpha}}, \bm{y}_{\alpha}),~
    \dot{\hat{\bm{\beta}}} = \bm{\Gamma}\text{Proj}(\hat{\bm{\beta}}, \bm{y}_{\beta})
\end{align}
where $\bm{\Gamma} \in \mathbb{R}^{6 \times 6}$ is a symmetric positive definite matrix, and the projection functions $\bm{y}_{\alpha} \in \mathbb{R}^{6}$ and $\bm{y}_{\beta} \in \mathbb{R}^{6}$ are given by:
\begin{align}\label{eq: proj_fun}
    \bm{y}_{\alpha}  = -{\bm{G}}^T \bm{P} \tilde{\bm{\eta}} \|\bm{\eta}\|, \quad
    \bm{y}_{\beta} = -{\bm{G}}^T \bm{P} \tilde{\bm{\eta}}.
\end{align}
The stability of the system using the adaptive control law \eqref{eq: control_law} and the adaptation laws \eqref{eq: adap_law} is proven in \cite{Nguyen2015}.
\section{Results} \label{sec: results}

To evaluate the proposed controller's performance, we present simulation and hardware experiment results. The simulation is conducted using the Gazebo simulator with ROS Noetic. We use the OCS2 package \cite{Farshidian2017OCS2:Systems} for solving the MPC problems, with inspiration from the open-source repository \cite{Legged_control} for the local MPC implementation. As shown in \figref{fig: block_diagram}, each control component runs at different frequencies: the global MPC at 100 Hz, the local MPC at 200 Hz, and the MRA-WBC at 800 Hz. The global MPC is given a time horizon of 5 seconds, while the local MPC has a 3-second horizon, allowing the global trajectory to be predicted slightly ahead of the local planner. Pinocchio \cite{pinocchioweb} is utilized for online computations of nonlinear dynamics and reference models within the MRA-WBC module. The terrain receiver generates the elevation map from point clouds and performs online segmentation for safe region extraction using open-source repositories \cite{elevation_mapping, plane_segmentation}. The supplementary video accompanying this paper showcases all the achieved results\footnote{\url{https://youtu.be/MuJY9rYxTO4}}.

\subsection{Obstacle Courses}
To evaluate the performance of our approach, we designed a scenario where two Unitree Aliengo robots face several challenges while carrying an unknown payload. The controller assumes the payload weighs 7 kg, but it actually weighs 10 kg, and we add random external forces in the range of \( -15 \leq F_{ext} \leq 15 \) to simulate a time-varying load. The obstacle course includes a large obstacle blocking half the path, a low vertical wall forcing the team to crawl, and a vertical bar requiring height adjustments, all on discrete stepping stones. As shown in \figref{fig: sim_fig}, the robots successfully navigate the obstacles by deviating from the reference trajectory, adjusting their height as needed, and compensating for the unknown payload mass while maintaining future safe foothold locations.

\subsection{Advantages of Hierarchical MPCs}
A common question is the benefit of having two separate global and local MPCs. What if we included collision avoidance constraints within the local MPC, having only a single MPC and allowed robots to share their optimal trajectories while considering holonomic constraints from the connecting rod? We tested this approach and found two key advantages of our method:

\subsubsection{Enhanced Trajectory Optimization}
The single MPC approach relies heavily on inter-robot communication, making it less proactive in trajectory optimization. In contrast, our centralized global planner delivers smoother trajectories and faster adjustments for obstacle and collision avoidance.

\subsubsection{Simplified Handling of Complex Scenarios} 
\begin{figure}[t!] {\includegraphics[width=1\linewidth]{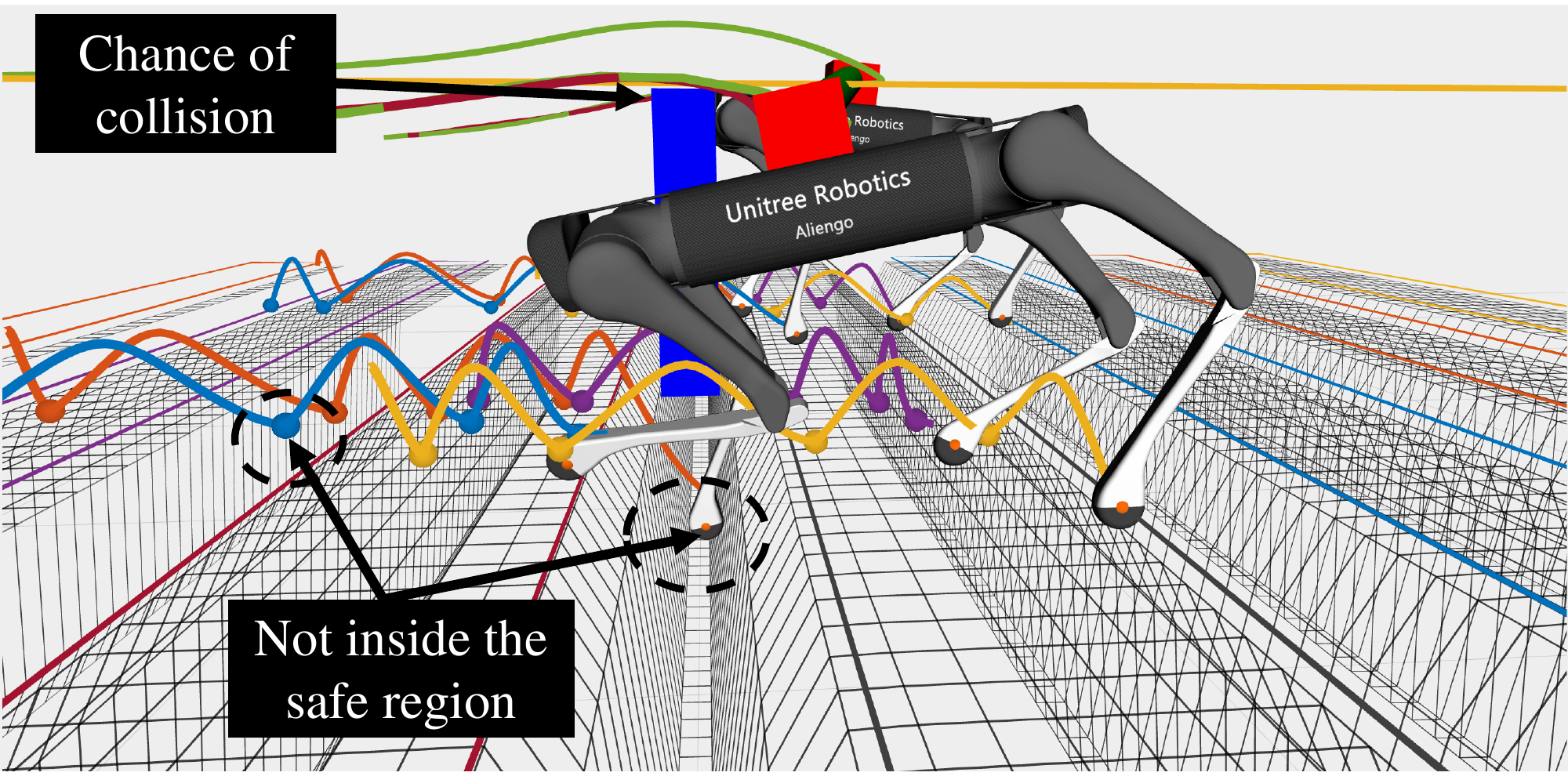}} \caption{\textbf{Single MPC Performance.} The figure shows the team navigating a stepping stone with a vertical bar using a single MPC planner. The MPC generates footholds outside the safe region and a high-risk trajectory, increasing the likelihood of obstacle collisions and locomotion failure.} 
\label{fig: sim_comparison} 
\vspace{-0.5cm} \end{figure}

In challenging situations, such as navigating discrete terrain while passing over the vertical bar, the single MPC approach struggles with obstacle avoidance and safe foothold planning. Our method simplifies parameter tuning and consistently performs well in both simulations and hardware tests across various scenarios. \figref{fig: sim_comparison} shows the single MPC method being tested as the team attempted to navigate over a vertical bar on a stepping stone. The figure highlights that the MPC generates footholds for the current and future steps outside the safe region. It also illustrates that the MPC struggles with generating an optimized trajectory for the payload and has a significant risk of colliding with obstacles, which is expected given the increasing complexity of the MPC problem.

\subsection{Hardware Experiment}
For the hardware experiment, we use a Unitree A1 and Go1 robot connected by a 1-meter rod, sharing a payload. The rod and container weigh 5 kg, which is known to the controller, while an additional 4 kg load inside is unknown. As the robots navigate, they encounter a box hanging 22 cm above the ground. Typically, robots are operating at a height of 30 cm. As the team approaches the box, the global MPC adjusts their height to pass underneath the obstacle safely and then guides them back to their nominal height. Snapshots of the experiment are shown in \figref{fig: hardware_fig}, and further details can be found in the supplemental video.
\begin{figure}[t!]
    \subfloat[]{\includegraphics[width=0.48\linewidth]{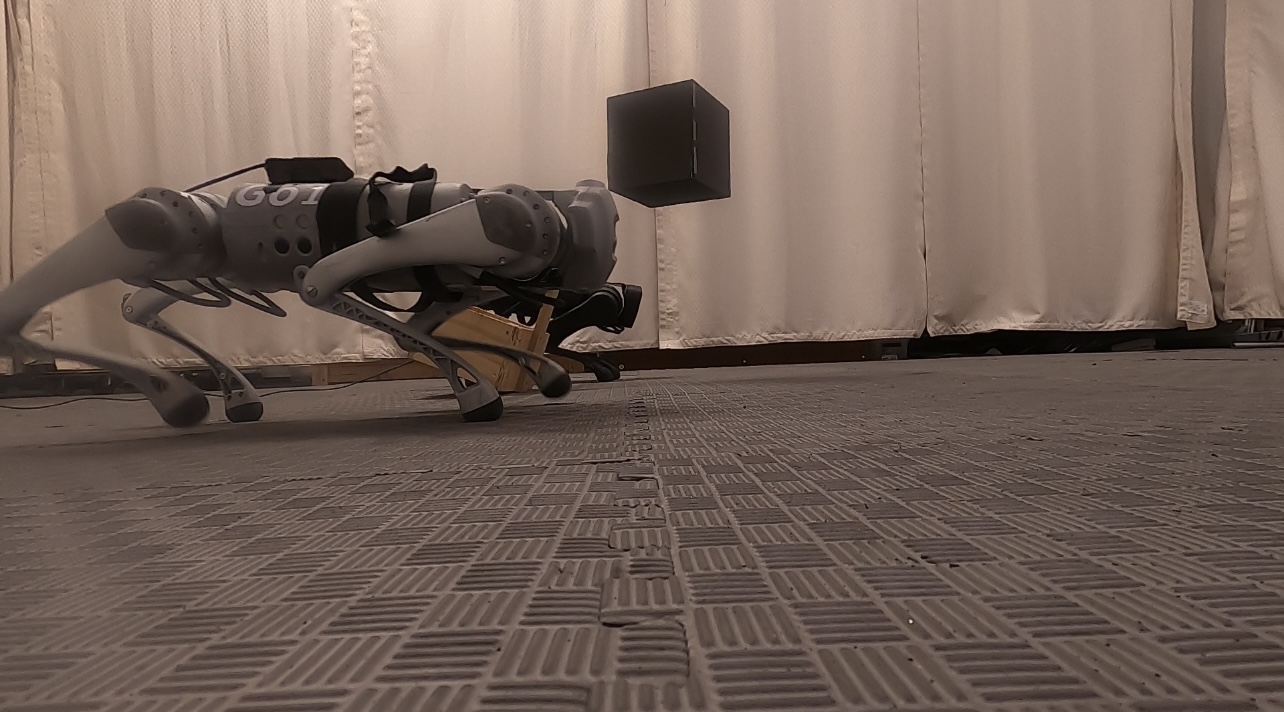}}
    \hfill
    \subfloat[]{\includegraphics[width=0.48\linewidth]{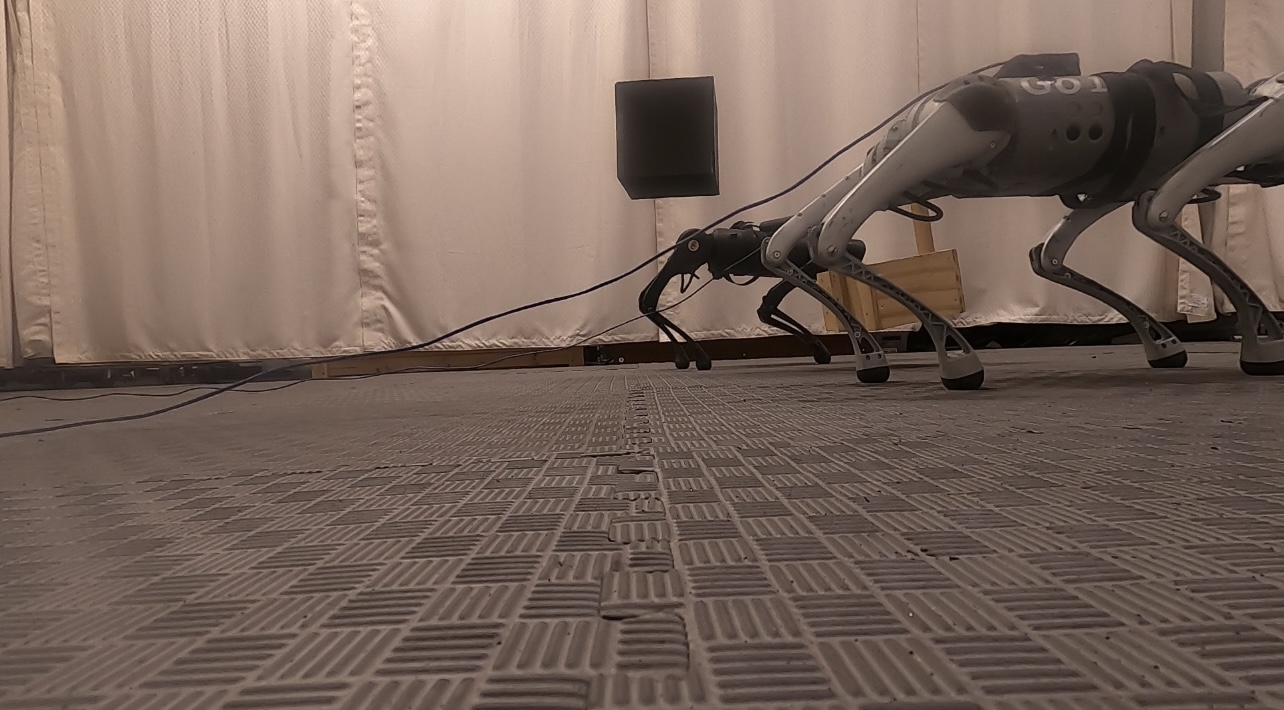}}
    \caption{\textbf{Hardware Experiment Snapshots.} (a) The robot team lowers their height to pass beneath the black box obstacle, and (b) they return to normal operating height after passing through the obstacle. All this coordination is achieved through the use of the global MPC.}
    \label{fig: hardware_fig}
    \vspace{-0.5cm}
\end{figure}

\section{Conclusion} \label{sec: conclusion}

This paper presented a safety-critical motion planning framework for collaborative legged loco-manipulation on discrete terrain. The approach integrates a global MPC for trajectory generation and obstacle avoidance, a local MPC for safe foothold planning and stable locomotion, and an MRA-WBC for adaptive tracking in the presence of payload uncertainties. We validated the method through simulations and hardware experiments, showing the robots' capability to navigate complex obstacle courses with unknown payloads on challenging terrains such as stepping stones.

Future work will extend the framework to more complex terrains with elevations by integrating bipedal robots and considering human-robot collaboration.

\balance
\bibliographystyle{IEEEtran}
\bibliography{references}

\end{document}